\documentclass[
11pt,%
tightenlines,%
twoside,%
onecolumn,%
nofloats,%
nobibnotes,%
nofootinbib,%
superscriptaddress,%
noshowpacs,%
centertags,
aps]%
{revtex4}
\usepackage{ljm}
\usepackage{tcolorbox}
\usepackage{tempora}
\tcbuselibrary{breakable}

\newcommand{\centered}[1]{\begin{tabular}{l} #1 \end{tabular}}

\newcommand{\thor}{\textsc{THoR}}

\newcommand{\rsneRu}{\rsne{}}
\newcommand{\rsneEn}{\rsne{}\textsuperscript{\texttt{en}}}

\newcommand{\rusentneCompetition}{RuSentNE-2023}
\newcommand{\rsne}{\rusentneCompetition{}}
\newcommand{\fpn}{$F1^{PN}$}
\newcommand{\fpnu}{$F1^{PN0}$}
\newcommand{\unknp}{N/A\textsubscript{\%}}
\newcommand{\unknOk}{$\cdot$}
\newcommand{\naclass}{UNK}

\newcommand{\chatgpt}{ChatGPT}
\newcommand{\gptTwo}{GPT-2}
\newcommand{\gptFour}{GPT-4}
\newcommand{\gptFourEleven}{GPT-4\textsubscript{1106-preview}}
\newcommand{\gptThree}{GPT-3.5}
\newcommand{\rutFive}[1]{ruT5\textsubscript{#1}}
\newcommand{\rugptThree}[1]{ruGPT-3\textsubscript{#1}}
\newcommand{\gptThreeSix}{GPT-3.5\textsubscript{turbo-0613}}
\newcommand{\gptThreeEleven}{GPT-3.5\textsubscript{turbo-1106}}
\newcommand{\mistralOne}{Mistral-Instruct-7B\textsubscript{v0.1}}
\newcommand{\mistralTwo}{Mistral-Instruct-7B\textsubscript{v0.2}}
\newcommand{\decilm}{DeciLM-7B}
\newcommand{\openai}{OpenAI}

\newcommand{\prompt}{PROMPT}
\newcommand{\promptVone}[1]{v1\textsubscript{#1}}
\newcommand{\promptVtwo}[1]{v2\textsubscript{#1}}

\begin{document}

\titlerunning{Large Language Models in Targeted Sentiment Analysis} 
\authorrunning{Nicolay Rusnachenko et al.} 

\title{Large Language Models in Targeted Sentiment Analysis for Russian}

\author{\firstname{N.}~\surname{Rusnachenko}}
\email[E-mail: ]{rusnicolay@gmail.com}
\affiliation{Newcastle Upon Tyne, England, United Kingdom}

\author{\firstname{A.}~\surname{Golubev}}
\email[E-mail: ]{antongolubev5@yandex.ru}
\affiliation{Lomonosov Moscow State University}

\author{\firstname{N.}~\surname{Loukachevitch}}
\email[E-mail: ]{louk_nat@mail.ru}
\affiliation{Lomonosov Moscow State University}
\affiliation{Research Computing Center Lomonosov Moscow State University}

\firstcollaboration{(Submitted by V.~V.~Voevodin)} 

\received{March 1, 2023} 

\begin{abstract} 
In this paper we investigate the use of decoder-based generative transformers for extracting sentiment towards the named entities in Russian news articles. 
We study sentiment analysis capabilities of instruction-tuned large language models (LLMs). 
We consider the dataset of
\rusentneCompetition{}
in our study.
The first group of experiments was aimed at the evaluation of zero-shot capabilities of LLMs with closed and open transparencies.
The second covers the fine-tuning of Flan-T5 using the "chain-of-thought" (CoT) three-hop reasoning framework (\thor{}).
We found that the results of the zero-shot approaches are similar to the results achieved by baseline fine-tuned encoder-based transformers (BERT\textsubscript{base}). 
Reasoning capabilities of the
fine-tuned Flan-T5 models with
\thor{}
achieve at least $5\%$ increment
with the base-size model compared to the results of the zero-shot experiment.
The best results of sentiment analysis  on \rsne{} were achieved by fine-tuned Flan-T5\textsubscript{xl}, which 
surpassed the results of previous state-of-the-art transformer-based classifiers. 
Our CoT application framework is publicly available: \url{https://github.com/nicolay-r/Reasoning-for-Sentiment-Analysis-Framework}
\end{abstract}

\subclass{12345, 54321} 


\maketitle

\section{Introduction}

In recent years, large language models (LLMs) based on the Transformer architecture have significantly changed the landscape of natural language processing (NLP). Such models are pre-trained on large volumes of unlabeled texts. The so-called instruction-tuned language models are further trained on large sets of instructions (prompts and correct answers). This pre-training made it possible to solve tasks without training (fine-tuning) models on target datasets in the so-called zero-shot or few-shot formats~\cite{brown2020,wei2021finetuned}. The zero-shot format is based solely on the formulation of a special prompt (question) for a model \cite{zhang2023stance,brown2020}.  The few-shot format comprises a prompt and several examples of correct answers \cite{brown2020}.
In addition, there are special prompts that contain instructions for reasoning, referred to as Chain-of-Thought (CoT) prompts \cite{wei2022chain}.


In sentiment analysis, the application of pre-trained language models has led to a significant improvement in the performance in various tasks. Sentiment analysis tasks can be divided into two main categories: general sentiment analysis ~\cite{turney-2002-thumbs}, and targeted sentiment analysis (TSA)~\cite{toledo2022multi}. General sentiment analysis aims to determine the overall sentiment of a text, while targeted sentiment analysis focuses on identifying the sentiment towards a specific entity~\cite{amigo2013overview,loukachevitch2015entity,golubev2023rusentne}, its characteristics (aspects)~\cite{pontiki2016semeval}, or controversial issues.

Large language models are primarily trained on English text collections and English datasets. Experiments with the models are also typically conducted for English. When applied to other languages, the results of large language models tend to be worse than for English. In this paper, we test several instruction-tuned large language models on a complicated task of targeted sentiment analysis of Russian texts. We experiment with models of different transparency such as <<closed models>> \chatgpt{} series (\gptThree{} and \gptFour{}) and <<open models>> limited by 7 billion (7B) parameters~\cite{mistral,decilm,gemma,li2023textbooks, chung2022scaling}. Such "small" open models can be applied in NLP tasks with limited computing resources (1 NVidia A100 card),  which is important for practical applications.  We use the dataset \rsne{} of Russian news texts annotated with sentiment towards the mentioned named entities.


\section{Related Work}

Evaluating language models in sentiment analysis, the authors of~\cite{zhang2023sentiment} compare the performance of LLMs such as \chatgpt{} (\gptThree{} and \gptFour{}), PaLM~\cite{chowdhery2022palm}, Flan-UL2~\cite{tay2023ul2}, and LLaMA~\cite{touvron2023llama} across
13 sentiment analysis tasks on 26 datasets and compare the results against small language models (SLMs)
trained on domain-specific datasets. The tasks under study include document- and sentence-level sentiment analysis, aspect-based sentiment analysis, and also such tasks as implicit sentiment, irony, hate speech detection, etc. With such a diverse range of tasks and models, the authors created a standardized template for prompts, containing the task name, definition, and desired output format.   The authors conclude that the zero-shot application of LLMs is already effective for simpler sentiment classification tasks, such as binary and trinary classification. However, for tasks that require structured sentiment outputs, such as aspect-based analysis, the performance of LLMs lags behind that of small models trained on specific domains: LMs often have a lower accuracy than fine-tuned ones.

In \cite{koto2024zeroshot} the authors evaluate large language models \gptThree{}, BLOOMZ, and XGLM in sentiment classification in 34 languages, including 6 high/medium-resource languages, 25 low-resource languages,
and 3 code-switching datasets. When prompting LLMs, six variants of prompts are used, and the obtained results are averaged. For high-resource languages, \gptThree{} achieves $77.5$\% by F-measure, for low-resource (African) languages shows 
$38.3$\% by F-measure in 3-way classification.

In~\cite{FeiAcl23THOR} authors illustrate the application of the CoT concept to extract implicit sentiments from users' reviews~\cite{pontiki-etal-2014-semeval,li2021learning}. According to the provided paradigms, the proposed Three-Hop-Reasoning (\thor{}) system could be treated as an emerged paradigm: task-agnostic schema of reasoning steps, with one-by-one components referred to sentiment analysis.
With these steps, authors aimed at extraction of <<aspects>>, with further <<opinion>> as atomic components 
for devising the final answer~\cite{FeiAcl23THOR}.
The authors conclude that the application of the fine-tuning process for instruction-tuned Flan-T5~\cite{chung2022scaling} results in models that surpass few-shot systems across publicly open systems~\cite{raffel2020exploring} and significantly outperform encoder-based classifiers~\cite{devlin-etal-2019-bert}. 
When it comes to limitations, authors conclude their beliefs on unleashing the full LLMs reasoning capabilities by applying \thor{} towards large enough models~\cite{FeiAcl23THOR}.

For Russian, there are several directions of related investigations:
aspect-based (SentiRuEval)~\cite{loukachevitch2015sentirueval} and
entity-oriented sentiment analysis
(\rusentneCompetition{})~\cite{golubev2020improving,golubev2023rusentne}.
The most recent advances in both tasks show that the
highest results are mainly achieved by
fine-tuned encoder-based classification language models~\cite{devlin-etal-2019-bert,liu2019roberta,smetanin2021deep,golubev2020improving}. The best results on \rusentneCompetition{} evaluation were obtained by ensembles of BERT-like encoder models \cite{golubev2023rusentne}.

Several recent work explores the application of generative models in sentiment analysis tasks.
The authors of \cite{chumakov2023generative} study generative models of the GPT family in the Aspect-Based Triplet Extraction Task (ASTE). 
The authors compare the 
few-shot strategies for the GPT-3 and \chatgpt{} (\gptThree{}) models 
with the fine-tuned Russian  
\rugptThree{small} and \rugptThree{large}
models, based on the \gptTwo{} architecture~\cite{radford2019language}. 
They found that a few-shot approach on \rugptThree{}
family models did not produce adequate results: fine-tuned \rugptThree{} models showed a significant improvement. 
The authors of \cite{moloshnikov2023named} adopt the T5 model~\cite{raffel2020exploring} and compare it with encoder-based approaches in the \rusentneCompetition{} evaluation. 
Two variants of the Russian-adapted models \rutFive{base} and \rutFive{large} were used. 
However, the best results obtained by \rutFive{large} are 7 percentage points lower than the top \rusentneCompetition{} submission~\cite{golubev2023rusentne}. 


\section{\rsne{} Evaluation and Dataset}

The \rsne{} dataset\footnote{Resources are publicly available:~\url{https://github.com/dialogue-evaluation/RuSentNE-evaluation}} is annotated with sentiment towards named entities in news texts.
News texts pose challenges for targeted sentiment analysis \cite{golubev2023rusentne}. 
At first, such texts may contain opinions conveyed by different subjects, including the author(s)’ attitudes, positions of cited sources, and relations of mentioned entities to each other. 
Secondly, some sentences contain several named entities with different sentiments, complicating the determination of sentiment towards each individual named entity. Thirdly, the majority of named entities in news texts are mentioned in neutral context, 
indicating a significant prevalence of the neutral class.
Last but not least, the significant
amount of sentiment in news texts is implicit in nature, 
primarily conveyed through entity actions. 

The source of the annotated sentiment in \rsne{} could be: 
(i)~an author, 
(ii)~another cited source, and 
(iii)~another entity mentioned in the text.
Sentiment annotation is a three-scale and has the following labels: positive, negative, and neutral.

For example, in the following sentence there is a negative sentiment to \textit{Hungary}  and neutral sentiment to \textit{German Economy Minister}. The source of the negative sentiment to \textit{Hungary} is the \textit{German Economy Minister}:
\begin{center}
\foreignlanguage{russian}{
Министр экономики Германии критикует Венгрию за налог на иностранных инвесторов.
}
 \\
(German Economy Minister criticizes Hungary for tax on foreign investors.)
\end{center}

Named entities of the following types represent objects of sentiment \cite{loukachevitch2023nerel}:

\begin{itemize}
\setlength\itemsep{0em}
    \item Person — physical person regarded as an individual,
    \item Organization — an organized group of people or company, 
    \item Country — a nation or a body of land with one government,
    \item Profession — jobs, positions in various organizations, and professional titles,
    \item Nationality — nouns denoting country citizens and adjectives corresponding to nations in contexts different from authority-related.
\end{itemize}

The distribution of entity types in the 
training (\texttt{train}), 
development (\texttt{dev}), and 
test (\texttt{test})
sets of the \rsne{} dataset is presented in  Table~\ref{tab:entity_distribution}.


\begin{table}[h]
    \centering
    \small
    \renewcommand{\arraystretch}{1} 
    \caption{Distribution of entity types in 
    training (\texttt{train}), 
    development (\texttt{dev})
    and test (\texttt{test})
    sets of the \rsne{} dataset}
    \begin{tabular}{l|cc|cc|cc}
    \hline
    & \multicolumn{2}{c}{\texttt{train}}
    & \multicolumn{2}{|c|}{\texttt{dev}} 
    & \multicolumn{2}{c}{\texttt{test}} \\
    \hline
     Entity type & \# & \% & \# & \% &\# & \%\\
    \hline
    Person &1934&29&857&30& 480&25\\
    Profession&1666&25&533&24&510&26\\
    Organization&1487&23&653&23&484&25\\
    Country&1274&19&686&19&363&19\\    
    Nationality&276&4&116&4&110&5\\
    \hline
    Total &6637&100&2845&100& 1947 & 100\\
    \hline
    \end{tabular}
    \label{tab:entity_distribution}
\end{table}

\section{Experimental Setup}

We experiment with the initial dataset \rsneRu{} and its English translation (\rsneEn{}). 
Since  most LLMs are trained on English data, 
we adopt GoogleTranslate\footnote{\url{https://pypi.org/project/googletrans/}} to automatically translate and compose 
\rsneEn{}.
To evaluate the predicted results, 
we follow the competition rules~\cite{golubev2023rusentne}
and adopt macro F1-measure ranged [0,~100] over:
(1)~positive and negative classes \fpn{},
(2)~all task classes \fpnu{}.

We investigate LLMs reasoning capabilities with the following modes:
(i) zero-shot, and (ii)~fine-tuning.
To conduct the experiment, our computational resources were limited by access to a 
single NVIDIA~A100~GPU~(40GB). 
The model training  and  inference code are implemented in Python-3.8.

\subsection{LLMs Zero-shot Experiments Setup}
\label{sec:zero-shot-setup}

Our setup covers a range of recently popular models that fall into two main categories: 
(i)~<<closed models>>, and 
(ii)~<<open models>>~\cite{mistral,decilm,gemma,li2023textbooks, chung2022scaling}.

In the case of closed models, the following  chat-based dialogue assistants were used:    
\begin{itemize}
    \item \gptFour{}, version 1106 preview;
    \item \gptThree{},  versions  1106, 0613.
\end{itemize}    
For the GPT models, we utilized a so-called system prompt, which is a  special message used to assign a role to the assistant. System prompts prescribe the  style and task for the chat-bot communication. We used the following system prompt:

\begin{tcolorbox}[breakable,colframe=gray,boxrule=2pt,arc=3.4pt,boxsep=-1mm]
\textbf{System Message}: You are an AI assistant skilled in natural language processing and sentiment analysis. Your task is to analyze text inputs to determine the underlying sentiment, whether it's positive, negative, or neutral. You should consider the nuances of language, including sarcasm, irony, and context. Your responses should include not only the sentiment classification but also a brief explanation of why a particular sentiment
was assigned, highlighting key words or phrases that influenced the decision.
\end{tcolorbox}

In the case of open models, we experiment with instruction-tuned versions.
The complete list of the assessed models is presented in Table~\ref{tab:llm-zs-experiments-list} and includes:
\begin{itemize}
        \item Mistral~\cite{mistral} -- grouped-query attention (GQA)~\cite{ainslie-etal-2023-gqa} for faster inference, coupled with sliding window attention~\cite{beltagy2020longformer} to effectively handle sequences of arbitrary length with a reduced inference cost.
    The authors adopt byte-fallback BPE tokenization~\cite{sennrich2016neural}.
    There are several versions of publicly available instruction-tuned models: \texttt{v0.1} and \texttt{v0.2}.
    The \texttt{v0.1} has the input context window size of 8K tokens\footnote{Technically is unlimited, with threshold originated by the 4K size of sliding window~\cite{beltagy2020longformer}}.
    In \texttt{v0.2} the related size has been increased up to 32K context\footnote{Our assumption that model is no longer adopts the sliding window attention~\cite{beltagy2020longformer}}. 
    For both versions, information on training data is not publicly available.
    
    \item DeciLM~\cite{decilm} -- is an auto-regressive language model using an optimized transformer decoder architecture that includes variable GQA~\cite{ainslie-etal-2023-gqa}.
    Information on fine-tuning and utilized datasets is publicly available.
    Model has been fine-tuned for instruction with LoRA~\cite{hu2021lora} on the SlimOrca~\cite{SlimOrca} dataset.
    The context window size is 8K tokens.
    
    \item Microsoft-Phi-2~\cite{li2023textbooks} -- trained using 
    the resource adopted in Phi-1.5~\cite{li2023textbooks}; augmented with a new data source that consists of various NLP synthetic texts and filtered websites (for safety and educational value). 
    Dataset represents combination of NLP synthetic data created by AOAI\footnote{\url{https://github.com/microsoft/sample-app-aoai-chatGPT}} \gptThree{} and filtered web data from Falcon RefinedWeb~\cite{refinedweb} and SlimPajama~\cite{cerebras2023slimpajama}, assessed by AOAI \gptFour{}.
    The context window size of the model is 2K tokens.
    For model training, authors utilize 96xA100-80G for 14 days. 
    
    \item Gemma~\cite{gemma} --
      Built from the same research and technology used to create the Gemini models~\cite{geminiteam2023gemini}. 
      Represent a text-to-text, decoder-only large language models, with architecture similar to LLaMA~\cite{touvron2023llama}. Available in English, with open access to instruction-tuned variants.  Represent well-suited for a variety of text generation tasks, including question answering, summarization, and reasoning. 
      The context window size is 8K tokens.
    \item Flan-T5~\cite{chung2022scaling} -- 
         Represent a specialized variant of the T5~\cite{t5paper}, initially proposed as the unified text-to-text transformer. 
         The concept of fine-tuning language models based on instructions (Flan)~\cite{chung2022scaling} 
         highlights the significant improvement across series of tasks, including: MMLU~\cite{hendrycks2021measuring}, 
         BBH~\cite{suzgun-etal-2023-challenging}, TyDiQA~\cite{clark-etal-2020-tydi}, MGSM~\cite{shi2022language}, open-ended generation.
         Flan-T5 trained with the encoder context length of 1K tokens.
\end{itemize}
Given the sentence ($X$) with the target entity mentioned in it ($t$),
we utilize prompting techniques of two types:
original version
(\promptVone{})~\cite{golubev2020improving} and 
precisely adapted
the revised version, dubbed as (\promptVtwo{}):
\begin{tcolorbox}[breakable,colframe=gray,boxrule=2pt,arc=3.4pt,boxsep=-1mm]
    \textbf{\promptVone{ru}}:
    \foreignlanguage{russian}{
    Какая оценка тональности в предложении $X$, по отношению к $t$? Выбери из трех вариантов: позитивная, негативная, нейтральная.
    }
    {\vspace{2mm}\noindent\rule{\textwidth}{0.4pt}}\\
    \textbf{\promptVone{en}}:
    What's the attitude of the sentence 
    $X$
    to the target $t$? 
    Select one from: positive, negative, neutral.
\end{tcolorbox}
\begin{tcolorbox}[breakable,colframe=gray,boxrule=2pt,arc=3.4pt,boxsep=-1mm]
    \textbf{\promptVtwo{ru}}:
    \foreignlanguage{russian}{
    Каково отношение автора или другого субъекта
    в предложении $X$ к $t$?
    Выбери из трех вариантов: позитивная, негативная, нейтральная
    }
    {\vspace{2mm}\noindent\rule{\textwidth}{0.4pt}}\\
    \textbf{\promptVtwo{en}}:
    What is the attitude of the author or another subject in the sentence $X$ to the target $t$? 
    Choose from: positive, negative, neutral.
\end{tcolorbox}


In the case of 
proprietary models, we 
adopt \openai{} API service to access the \openai{} models. To reduce the charging cost for the tokens, we limit the result output for \gptThree{} and \gptFour{} models by using the response threshold of $75$ tokens.
We augment the initial prompt with the suffix that requires short answers as follows:
<<Create a very short summary that uses $50$ \texttt{completion\_tokens} or less.>> 
For LLMs that can be downloaded for local use, in this paper, we experiment with models 
whose size does not exceed 7B parameters. 
We utilized \texttt{transformers} API~\cite{wolf-etal-2020-transformers} to conduct experiments on rented servers. 
In all models for zero-shot experiments, the value of the temperature parameter was chosen as $0.1$.

\begin{table}[!t]
    \centering
    \renewcommand{\arraystretch}{1} 
    \caption{List of LLMs utilized in Zero-shot experiments, separated on <<closed models>> with access via OpenAI API (GPT) and <<open models>>~\cite{mistral,decilm,gemma,li2023textbooks, chung2022scaling} which size does not exceed 7 billion parameters}
    \begin{tabular}{p{1.9cm}lll}
    \hline
    Models & Versions & Params & Reference \\
    \hline
        \gptFour{}  & 1106-preview & 1.76T\footnote{Non-disclosured by OpenAI; according to the other sources, \gptFour{} yields of eight models 220B-sized parameters each connected by a Mixture of Experts (MoE)~\cite{gpt4paramsCount} ($\approx$1.76T params)} & \texttt{gpt-4-1106-preview} (OpenAI API)\\
        \gptThree{} & 1106 & 175B\footnote{Non-disclosed by OpenAI, our assumption that the architecture is referred to GPT-3, size of 175B params~\cite{brown2020}}  & \texttt{gpt-3.5-turbo-1106} (OpenAI API)\\
                    & 0613 & 175B\textsuperscript{$b$}  & \texttt{gpt-3.5-turbo-0613} (OpenAI API)\\
        \hline
        Mistral~\cite{mistral} & v0.1 & 7B & \url{https://huggingface.co/mistralai/Mistral-7B-Instruct-v0.1} \\
                               & v0.2 & 7B &
                               \url{https://huggingface.co/mistralai/Mistral-7B-Instruct-v0.2}
                               \\
        \cline{2-4}
        DeciLM:~\cite{decilm} & & 7B &\url{https://huggingface.co/Deci/DeciLM-7B-instruct}\\
        \cline{2-4}
        Microsoft-Phi-2~\cite{li2023textbooks} & & 2.7B & \url{https://huggingface.co/microsoft/phi-2}\\
        \cline{2-4}
        Gemma:~\cite{gemma} & Instructive & 7B\footnote{The actual and non-official amount of hidden parameters for Gemma-7B-IT model is 9B}& \url{https://huggingface.co/google/gemma-7b-it}\\
                           & Instructive & 2B&
                           \url{https://huggingface.co/google/gemma-2b-it}
                           \\
        \cline{2-4}
        Flan-T5:~\cite{chung2022scaling} & XL    & 3B  & \url{https://huggingface.co/google/flan-t5-xl}\\
                                         & Large & 750M&
                                         \url{https://huggingface.co/google/flan-t5-large}\\
                                         & Base  & 250M& 
                                        \url{https://huggingface.co/google/flan-t5-base} 
                                         \\
    \hline
    \end{tabular}
    \label{tab:llm-zs-experiments-list}
\end{table}

To assess the inferred textual responses, 
in this study we adopt a universal annotation answer mapping strategy that involves the 
application of class-specific textual lower-cased templates for classes (positive, negative, and neutral), 
separately declared for each language.
We use these templates during a sequential occurrence check in lower-cased output in the following order: (1)~neutral, (2)~positive, (3)~negative. 
In the case of absence of any class, the <<\naclass{}>> placeholder 
(counted as <<neutral>>) 
was considered. 
We use 
<<\texttt{positive}>>,
<<\texttt{negative}>>,
<<\texttt{neutral}>>
templates
for \rsneEn{}, 
and
\foreignlanguage{russian}{
<<\texttt{позитив}>>} (positive),
\foreignlanguage{russian}{
<<\texttt{негатив}>>} (negative),
\foreignlanguage{russian}{
<<\texttt{нейтрал}>>} (neutral)
templates
for experiments on \rsneRu{} dataset. 
We perform final checks involving regular expressions to guarantee that the final answers are not prefixed with the initial prompt.


\subsection{LLMs Fine-tuning  Setup}
To experiment with the fine-tuning, we adopt encoder-decoder style instruction-tuned Flan-T5\footnote{\scriptsize\url{https://huggingface.co/docs/transformers/en/model_doc/flan-t5}} 
as our backbone large language model for the proposed methodology for texts written in English.
In this paper we experiment with the following fine-tuning techniques: 
\begin{itemize}
    \item \prompt{} -- \textsc{prompt}-tuning with the \promptVone{en}
    version of the prompt; 
    \item \thor{} -- Three-Hop-Reasoning~\cite{FeiAcl23THOR} technique, which can  be considered as an emerged paradigm: task-agnostic schema of reasoning steps, with one-by-one components  referred to sentiment analysis. The experiments with three-hop reasoning are resource and time-intensive, so we currently  study this technique only for one LLMs family.
\end{itemize}

\begin{figure}[!t]
    \includegraphics[width=\textwidth]{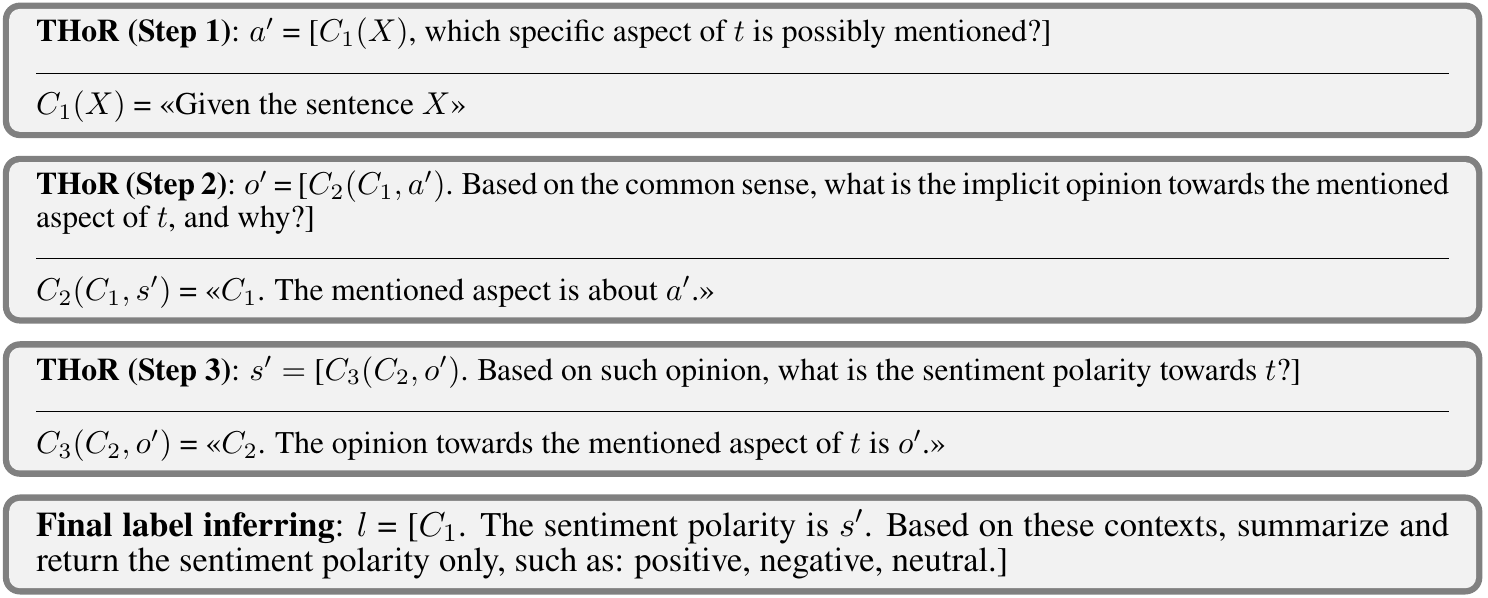}
    \caption{Inferring sentiment $s`$ using CoT three-hop reasoning framework (\thor{}), including <<\textit{final label inferring}>> to answer one of the task classes~\cite{FeiAcl23THOR}}
    \label{fig:thor-cot}
\end{figure}

Next, we cover \thor{} in greater details.
With $C_{i}, i\in \overline{1..3}$ we denote the prompts that wrap the content in the input context.
Given the sentence ($X$) with target entity mentioned in it ($t$), 
Figure~\ref{fig:thor-cot} illustrates the
initial application of the three-step approach~\cite{FeiAcl23THOR} 
for inferring the sentiment $s'$.
According to Table~\ref{fig:thor-cot}, the result $a'$ could be interpret as $a' = argmax~p(a|X,t)$,
opinion $o'$ as $o' = argmax~p(o|X,t,a'_{1})$, and
the final answer $s'$ noted as: $s'= argmax~p(e|X,t,s',o')$.
To guarantee the correctness of the final answer ($s'$),
we use the following prompt message to inferring the sentiment label~$l$~(Figure~\ref{fig:thor-cot}, bottom).

We experiment with: 
250M (base),
750M (large),
and 
3B (XL)
versions.
We conduct experiments  only for English-translated \rsneEn{} dataset  due to the pre-training specifics of Flan-T5 model.
For mapping the Flan-T5 output towards 
task classes, 
we seek for the exact string from the set of textual task labels:
<<\texttt{positive}>>, 
<<\texttt{negative}>>,
and <<\texttt{neutral}>>.
When it comes to fine-tuning parameters setup, we follow the settings proposed by authors of \thor{} framework~\cite{FeiAcl23THOR}.
In particular, we use AdamW~\cite{loshchilov2018decoupled}
optimizer 
with learning rate 
$2\cdot 10^{-4}$
and \textsc{batch-size} of $32$.
To infer the answers, the maximal temperature of
$1.0$ was considered.

\section{Results and Discussion}

Table~\ref{tab:zeroshot-models-results} shows the 
zero-shot prompting results
on the \texttt{test} set of \rsne{}
across various LLMs,
separately for 
the 
original 
and 
translated texts (\rsneEn{}).
The Table~\ref{tab:zeroshot-models-results} contains two main measures for evaluation (\fpn{}, \fpnu{}), accompanied by \textit{no-answer rate} (\unknp{}). In general, all models were capable of handling instructions generated using \promptVone{} and \promptVtwo{} prompts  for entity-oriented  sentiment analysis of initial and translated versions of the  
\rsne{}
dataset.



It can be seen from Table~\ref{tab:zeroshot-models-results} that such models that support Russian language are tend to perform 
worse with texts from \rsneRu{} than for its translated variant \rsneEn{}. There are a few models for which the proportion of \unknp{} answers (Table~\ref{tab:zeroshot-models-results}) is significantly higher than for other models. The  Microsoft-Phi-2 model with both versions of prompts returns the prompt along with its response. Since the full response is limited by the \texttt{max\_token} value, it is truncated, and we do not receive information about sentiment classification from the model. The average length of a response for examples for which the model did not provide an answer is 465 characters, for the rest it equals 288. \decilm{} and Gemma-2B models with \promptVtwo{} prompt configuration generate garbage answers in the form of data scraps from the prompt or clearly indicate that they cannot determine the sentiment polarity (<<\texttt{...so i cannot answer this question from the provided context.}>>).





According to \fpn{}, in the case of proprietary \openai{} models   
we see the similar performance order of the models' results 
irrespective of the prompts language source.
In particular, the highest results on \rsneEn{}
were achieved by \gptFour{}
(\fpn=~$54.53$)
, followed by 
\gptThreeSix{} 
(\fpn=~$49.22$) 
and 
\gptThreeEleven{} 
(\fpn=~$47.87$).\footnote{
We believe that the reason of the worse behavior of the \gptThreeEleven{} against the previous \gptThreeSix{} is caused by a higher tolerance level of the results responses.
}
Switching to the original \rsne{} results in 
10\% and 7\% decrease in result
by \fpn{} for \gptFour{} and \gptThreeSix{} respectively.
In turn, the only multilingual \mistralTwo{} illustrates a performance comparable to \gptThreeSix{} and outperforms \gptThreeEleven{} by $\approx$19-25\%.
Also, we found \mistralTwo{} as more sensitive to prompts on original \rsne{} texts, rather OpenAI models (see last row, Table~\ref{tab:zeroshot-models-results}).

Most open LLMs 
are able to follow English instructions. 
For \rsneEn{}, we see that the best performing models are Mistral~\cite{mistral} 
models ($45.69\leq$~\fpn{}~$\leq49.56$), 
and \decilm{}~\cite{decilm} as the closest competitor alternative 
(\fpn{}=~$42.73$).
The remaining families of 
Microsoft-Phi-2 and 
Flan-T5 series demonstrate the gap in results
($32.64\leq$~\fpn{}~$\leq37.26$). 
Through the entire range of open models,
Mistral~\cite{mistral} is the only one capable of handling \rsne{} in Russian.  
The official release \mistralTwo{} was better suited for non-English languages.

\begin{table}[!t]
    \centering
    \small
    \renewcommand{\arraystretch}{1} 

    \caption{Results of the LLMs application in zero-shot mode  
    for the 
    \texttt{test} part of the \rsne{} dataset,
    separately for original texts
    and automatically translated in English 
    (\rsneEn{}); for <<\unknp{}>> column values (no-answer rate),
    <<\unknOk{}>> denotes cases when the amount of unknown answers does not exceed 1\%; top results per each version of the dataset are bolded
    }
    \begin{tabular}{lr|ccc|ccc}
    \hline
    Model            & & \centered{\fpn{}} & \centered{\fpnu{}} & \unknp{} & \centered{\fpn{}} & \centered{\fpnu{}} & \unknp{} \\
    \hline
    \multicolumn{6}{l}{\rsneEn{} \textit{[Translated Texts into English]}} \\
    \hline 
    Prompt Type & & \multicolumn{3}{c}{\promptVone{en}}  & \multicolumn{3}{|c}{\promptVtwo{en}}          \\
    \hline
    \gptFourEleven{}      &  & \textbf{54.43}&\textbf{63.44} &\unknOk{} &\textbf{54.59}  &\textbf{64.32} &\unknOk{}\\
    \gptThreeSix{}  &  & 49.22& 59.51& \unknOk{}&  51.79 &61.38 &\unknOk{} \\
    \gptThreeEleven{} &  & 47.87& 54.62& \unknOk{}& 47.04 &53.19 &\unknOk{} \\
    \hline
    \mistralOne{}  &   & 49.56& 58.86& \unknOk{} &49.46  &58.51 &\unknOk{} \\
    \mistralTwo{}  &  & 45.69& 57.16& \unknOk{}&   44.82 &56.04 &\unknOk{} \\
    \cline{1-8}
    \decilm{}      &   & 42.73& 49.88& \unknOk{}&  43.85  &53.65 & 1.44 \\
    \cline{1-8}
    Microsoft-Phi-2&  & 37.26& 31.91& 5.55&40.95  &42.77 &3.13 \\
    \cline{1-8}
    Gemma-7B-IT&  & 40.58&45.94 &\unknOk{}& 40.96 &44.63 & \unknOk{}\\
    Gemma-2B-IT&  & 18.70& 39.51& \unknOk{}& 31.75 & 45.96& 2.62\\
    \cline{1-8}
    
    Flan-T5\textsubscript{xl}       &  & 35.35 & 31.51 & \unknOk{} & 48.14 & 57.33 &\unknOk{} \\
    Flan-T5\textsubscript{large}    &  & 34.86 & 23.34 & \unknOk{} & 36.05 & 24.27 &\unknOk{} \\ 
    Flan-T5\textsubscript{base}     &  & 32.64 & 21.81 & \unknOk{} & 31.05 & 20.84 &\unknOk{} \\ 
    \hline
    \hline
    \multicolumn{8}{l}{\rsneRu{} \textit{[Original texts written in Russian]}} \\
    \hline
    Prompt Type & & \multicolumn{3}{c}{\promptVone{ru}}  & \multicolumn{3}{|c}{\promptVtwo{ru}}          \\
    \hline
    \gptFourEleven{} &  & \textbf{49.44}& \textbf{58.74}& \unknOk{}& \textbf{48.04} & \textbf{60.55} &\unknOk{} \\
    \gptThreeSix{} &  & 45.97& 56.10& \unknOk{}&  45.85 &57.36 &\unknOk{} \\
    \gptThreeEleven{} &  & 38.95& 45.93& \unknOk{}& 35.07  &48.53 &\unknOk{} \\  
    \hline
    \mistralTwo{} &  & 48.71& 57.10& \unknOk{} & 42.60&48.05 &\unknOk{} \\
    \hline 
    \end{tabular}
    \label{tab:zeroshot-models-results}
\end{table}

\begin{figure}[!t]
    \centering
    \includegraphics[width=0.45\textwidth]{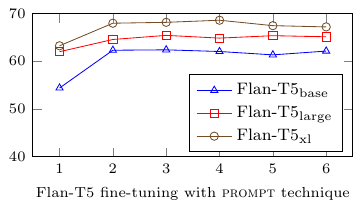}~
    \includegraphics[width=0.45\textwidth]{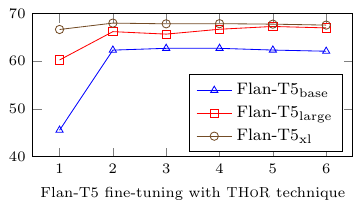}
    \caption{
    Analysis of the Flan-T5 models results on 
    \rsneEn{}
    \texttt{dev} 
    per each epoch (horizontal axis)
    by $F_1(PN)$ (vertical axis)
    during fine-tuning with \prompt{} (left)
    and \thor{} technique (right) per different sizes
    }
    \label{fig:finetuning-flant5-dev}
\end{figure}

From experiments with model fine-tuning,
we found that training Flan-T5 for 2-3 epochs
both for \prompt{} and \thor{}  techniques on  \rsneEn{} is sufficient.
Figure~\ref{fig:finetuning-flant5-dev} illustrates the
evaluation statistics of Flan-T5 models on \texttt{dev} dataset per each epoch of the training process (6 epochs in total), separately per each training technique.
We found that training for 2-4 epochs is sufficient
to prevent the model from overfitting.
For the final evaluation on the \texttt{test} set, checkpoints with the highest results on \texttt{dev} set were considered.
Table~\ref{tab:cot-ft-results} illustrates the results of the fine-tuning instruction-tuned Flan-T5 models.
We first compare and discuss the obtained results in comparison with zero-shot approaches, followed by comparison of the different fine-tuning techniques. 

Fine-tuning Flan-T5 on
\rsneEn{}
training data results in models that outperform all zero-shot approaches.
Since the Flan-T5 only supports texts written in English, we experiment with a \rsneEn{} dataset.
Comparing the results with those in Table~\ref{tab:zeroshot-models-results}, we see that the fine-tuned versions of the Flan-T5 models of all sizes significantly outperformed their corresponding zero-shot counterparts. The borderline for the Flan-T5\textsubscript{base} version is 
\fpn{}$=59.75$, 
which is 
$+9.9\%$ 
higher than the top result of
\gptFour{}
applied for \rsneEn{}
(\fpn{}$=54.36$, see Table~\ref{tab:zeroshot-models-results}).

Comparing the results of different fine-tuning techniques, we find that using
\thor{}
results in more stable performance across all model sizes. 
The exceptional case of the \texttt{xl} sized model, 
with which we see similar results on \texttt{dev} between different fine-tuning techniques with 
higher gap on \texttt{test} set.
According to the findings  in~\cite{wei2022chain}, 
the obtained results illustrate the alignment of the ideas that were investigated in
\thor{}
application for implicit sentiment analysis~\cite{li2021learning}.
Analyzing results on \texttt{test} part,
the finetuned model with \thor{} technique on
shows $+4.2$\% increment by \fpn{}
once switching from base-size to large-sized, 
and extra $+4.4$\% by \fpn{} 
with the XL-sized over large.
The highest achieved result is \fpn{}$=68.20$ 
which outperforms  the best \rsne{} results based on the transformer encoder ensemble~\cite{ai2023half}.

\begin{table}[!htp]
\renewcommand{\arraystretch}{1} 
\caption{Results of the Flan-T5 fine-tuning  
with (i)~\prompt{} and (ii)~\thor{} techniques for 
\rsneEn{} and results of the \thor{} in zero-shot mode for the comparison; top results per each column are bolded 
}
\begin{tabular}{l|l|cc|cc}
\hline
\multirow{3}{*}{Model} 
& \multirow{3}{*}{Technique} 
& \multicolumn{4}{c}{\rsneEn{}} \\
\cline{3-6} 
&& 
\multicolumn{2}{c}{\texttt{dev}} & 
\multicolumn{2}{c}{\texttt{test}} \\
\cline{3-6}
                       &         & \centered{\fpn{}} & \centered{\fpnu{}} &  \centered{\fpn{}} & \centered{\fpnu{}} \\
\hline
\multicolumn{6}{l}{Flan-T5 fine-tuning results}\\
\hline
Flan-T5\textsubscript{xl}             & \thor{}    & 68.02     & 74.82     & 65.09           & 72.45                      \\
Flan-T5\textsubscript{xl}             & \prompt{} (\promptVone{en})  & 68.62     & 75.69     & \textbf{68.20}            & \textbf{75.29}             \\
\hline
Flan-T5\textsubscript{large}          & \thor{}    & 67.31     & 74.67     & 62.29           & 70.70                      \\
Flan-T5\textsubscript{large}          & \prompt{} (\promptVone{en}) & 65.83     & 73.71     & 60.80           & 69.79                      \\
\hline
Flan-T5\textsubscript{base}           & \thor{}    & 62.72     & 70.70     & 59.75           & 68.02                      \\
Flan-T5\textsubscript{base}           & \prompt{} (\promptVone{en}) & 62.40     & 70.68     & 57.01           & 66.89                      \\
\hline
\multicolumn{6}{l}{Zero-shot results}\\
\hline
\gptFourEleven{}\footnote{Due to the high \gptFourEleven{} API cost, results were inferred for the \texttt{test} part only.} &  \thor{}              & --  &  --    & 50.13 & 55.93  \\
\gptThreeSix{} &  \thor{}                & 43.41 & 46.14 & 44.50 & 48.17 \\
\gptThreeEleven{} &  \thor{}             & 40.85 & 40.04 & 42.58 & 42.18 \\
Flan-T5\textsubscript{xl} &  \thor{}     & 38.30 & 32.12 & 38.58& 33.55\\
Flan-T5\textsubscript{large}&   \thor{}  & 34.66 & 23.10 & 34.69& 23.13\\
Flan-T5\textsubscript{base}&   \thor{}   & 33.93 & 22.79 & 33.88& 23.00\\
\hline
\hline
Best RuSentNE-2023 \cite{ai2023half} &Ensemble of encoders& \textbf{70.94} & \textbf{77.63} &66.67&74.11\\
\hline
\end{tabular}
\label{tab:cot-ft-results}
\end{table}




\section{Error Analysis}

For the error analysis, we selected examples from the \rsne{} test set where most models' predictions did not align with human annotations. The following main types of discrepancies between models' predictions and manual annotations are found (denoted as~<<E1>>, <<E2>>, and <<E3>>):
\vspace{2mm}
\\
\textbf{E1.} A sentence mentions the positive author's attitude to the target person and some negative event with this person (trauma, death).  The annotators treat such examples as positive for the person because in most cases traumas or death do no influence on existing positive attitude. 
Models in zero-shot mode answer on such examples by choosing the negative sentiment to the target person due to “negative effect” context. In turn, fine-tuned models annotate most of such examples correctly. For instance, in the following example we can see that the author has a positive attitude towards Chuck Berry despite the incident described.

\begin{center}
\foreignlanguage{russian}{
Легендарный Чак Берри потерял сознание на концерте.
}
 \\
(Legendary musician Chuck Berry fainted during a concert in Chicago.)
\end{center}
\textbf{E2.}	
A sentence mentions several entities, while the negative sentiment is directed only at one of these entities.   Models cannot distinguish the correct object of this negative attitude. For example:

\begin{center}
\foreignlanguage{russian}{
Юлия же в свою очередь обвиняет бывшего супруга в том, что он не выполняет решение суда, и уже объявлен в федеральный розыск.	
}
 \\
(Yulia, in turn, accuses her ex-husband of not complying with the court decision and has already been put on the federal wanted list.)
\end{center}
In this case, the models predict a negative attitude towards Yulia, but in fact Yulia has a negative opinion of her husband.
\vspace{2mm}
\\
\textbf{E3.} Similar to the E2. A sentence with evident negative sentiment mentions a single entity, but sentiment is directed to some out-of-sentence entity. Almost all models in zero-shot mode predict a negative sentiment to the mentioned entity. In the following example, the models in zero-shot mode answer by choosing negative sentiment to America, but in fact the negative opinion is towards the "situation".
\begin{center}
\foreignlanguage{russian}{
Ситуация, однако, не может нас не беспокоить -- объем экспорта в Америку упал на 8\%.}
\\
(The situation, however, cannot but worry us -- the volume of exports to America fell by 8\%.)
\end{center}

\section{Conclusion}

In this paper, we investigated the application of large language models (LLM) in targeted sentiment analysis within news texts.
We follow the 
\rusentneCompetition{}
competition 
both for experiments with LLM and evaluation.
The 
\rusentneCompetition{}
evaluation was aimed at extracting sentiment towards the objects annotated  in Russian-language  sentences.
In experiments, we used the \rsneRu{} 
 dataset as well as its English automatically translated version (\rsneEn{}).

 LLM-based sentiment analysis  was investigated in several directions, which cover the influence of aspects such as: 
(i)~the language of the source texts (Russian or English), 
(ii)~variants of prompts, and
(iii)~the effect of fine-tuning, including the application of the Chain-of-Thought technique.
We have discovered the significance of the content translation, since most LLM models demonstrate  reasonable comprehension  capabilities primarily in English.
 Zero-shot approaches achieve results comparable to fine-tuned BERT-based
\rusentneCompetition{}
competition baselines.
Experiments with LLM fine-tuning (Flan-T5)
have shown the improvement in $\approx10\%$ performance by $F_1(PN)$
for all considered Flan-T5 models.
The highest results were obtained by fine-tuned xl-sized Flan-T5\textsubscript{xl} model ($3$B+ parameters), which are currently   the best   results achieved on the \rsne{} test set.

In further, we aim to continue  experimenting with the Chain-of-Thought in the following directions:
(i)~reasoning revision techniques using extensive resources of auxiliary information,
(ii)~parameter-efficient tuning for larger models.

\section*{Acknowledgments}
The work is supported by the Russian Science Foundation (grant No. 21-71-30003).

\bibliography{main}{}

\end{document}